\documentclass{article}

\PassOptionsToPackage{numbers, compress}{natbib}

 \usepackage[preprint]{nips_2018}

\usepackage[utf8]{inputenc} 
\usepackage[T1]{fontenc}    
\usepackage{hyperref}       
\usepackage{url}            
\usepackage{booktabs}       
\usepackage{amsfonts}       
\usepackage{nicefrac}       
\usepackage{microtype}      

\usepackage{tablefootnote}
\usepackage{footnote}
\usepackage{color}
\usepackage{url}
\usepackage{subcaption}
\usepackage{caption}
\usepackage{verbatim}
\usepackage{enumitem}
\usepackage{wrapfig}
\usepackage[capitalise]{cleveref}
\usepackage{seqsplit}

\usepackage{graphicx}

\newcommand{\rulesep}{\unskip\ \vrule\ }

\title{Improving the Resolution of CNN Feature Maps Efficiently with Multisampling}

\author{
  Shayan Sadigh\thanks{\url{https://github.com/shayanpersonal}.} \\
  University of California, Santa Barbara \\
  ssadigh.contact@gmail.com
  \And
  Pradeep Sen\thanks{\url{http://www.ece.ucsb.edu/\~psen}} \\
  University of California, Santa Barbara \\
  psen@ece.ucsb.edu
}

\begin{document}

\maketitle

\begin{abstract}
We describe a new class of subsampling techniques for CNNs, termed multisampling, that significantly increases the amount of information kept by feature maps through subsampling layers. One version of our method, which we call checkered subsampling, significantly improves the accuracy of state-of-the-art architectures such as DenseNet and ResNet without any additional parameters and, remarkably, improves the accuracy of certain pretrained ImageNet models \textit{without any training or fine-tuning}. We glean possible insight into the nature of data augmentations and demonstrate experimentally that coarse feature maps are bottlenecking the performance of neural networks in image classification.
\end{abstract}

\section{Introduction}

Many applications of deep convolutional neural networks (CNNs), such as image classification and semantic segmentation, require the network to be able to capture the presence of large objects or features over the input. Most architectures add subsampling layers, such as max-poooling layers or convolutional layers where stride > 1, periodically throughout the network to reduce the spatial dimension lengths of feature maps and increase the receptive field of proceeding neurons. By reducing resolution, subsampling layers also reduce the computational complexity of deep layers.

Unfortunately, subsampling layers lose a significant amount of spatial information that could be highly informative to proceeding layers. Tasks that rely on fine-grained spatial information to generate accurate outputs, such as semantic segmentation, often attempt to address this issue with dilated convolutions \cite{dilated}, which increase the receptive field of convolutions without subsampling. However, subsampling is critical to reducing the computational complexity of deep layers, so these models still require the use of regular subsampling layers to make deep layers tractable \cite{deeplab, deeplabv3,deeplabv3plus2018, semsegreview}. Additionally, the choice of when and where to increase dilation over applying subsampling is fairly arbitrary and adds extra engineering overhead to CNN design. 

Furthermore, outside of fine-grained tasks such as semantic segmentation, there has been little progress in improving the resolution of feature maps. Nearly all image classification models use very coarse final feature maps (common sizes include $7\times$7 and $8\times$8) \cite{pyramidnet, resnet, squeezeandexcite, densenet} which we claim bottlenecks their accuracy. To address this problem, we rethink the representation of feature maps and make the following contributions:

\begin{enumerate}
\item We introduce \textbf{multisampling}, a technique to increase the number of samples taken from feature maps at subsampling layers, and thereby preserves more information for processing in late stages of the network. Traditional subsampling layers and increasingly dilated layers can be viewed as opposite, extreme types of multisampling.
\item We describe \textbf{checkered subsampling}, an instance of multisampling designed for 2D CNNs that use subsampling layers with a stride length of 2. Checkered subsampling, named for the checkerboard patterns it produces, preserves 50\% of the spatial resolution of the input feature map as opposed to the 25\% preserved with traditional subsampling. Repeated applications of checkered subsampling produce a denser, better-distributed sampling of the input compared to traditional subsampling layers.
\item We extend feature maps with a \textbf{submap dimension} to store features produced by multisampling. A feature map can be represented by different \textbf{feature submaps} stored across the submap dimension. Operations can be applied across the submap dimension with 3D layers.
\end{enumerate}

We refer to a CNN that use checkered subsampling as a \textbf{checkered CNN} or \textbf{CCNN}. Many common architectures can be easily converted into CCNNs and show significantly better accuracy than their traditional CNN counterparts across the board. Some pretrained ImageNet models can be converted to CCNNs and immediately show improved accuracy without any training.

Checkered subsampling maintains the core benefits of traditional subsampling while significantly increasing spatial resolution. That is, the spatial dimension lengths and spatial resolution of feature maps are both reduced with each checkered subsampling layer, enabling the network to learn large-scale features and, importantly, \textit{reducing the computational costs of deep layers} (in contrast to dilation, which is computationally expensive to implement). Our complexity over traditional subsampling layers per layer is $\mathcal{O}(n)$, where $n$ is the number of samples taken by multisampling (in traditional subsampling layers, \(n\) is always 1). Our technique is simple to implement in deep learning frameworks such as PyTorch, and popular CNN architectures such as ResNet and DenseNet can take advantage of checkered subsampling with minimal code changes.

\section{Related work}
\textbf{Dilated convolutions} \cite{dilated}, also referred to as \`{a}-trous convolutions \cite{deeplab}, are commonly used to increase the receptive field of kernels without subsampling. Dilated convolutions are similar to multisampling in that they are both techniques for preserving the spatial resolution of feature maps while increasing the receptive field of neurons. However, a key drawback of dilated convolutions is exactly that \textit{they do not perform any subsampling}, which is important for reducing the computational complexity of deep layers. Thus, models that use dilation still rely on regular subsampling layers. Dai et al. \cite{deformable} extend dilation with \textbf{deformable convolutions}. 3D kernels that look across the submap dimension can replicate some of the effects of deformable convolutions due to the semi-structured nature of the submap dimension.

The similarly named \textbf{multipooling} \cite{multipool} is used to improve the run-time performance of patch-based CNN methods by avoiding the redundant processing of overlapping patches. Multipooling shares algorithmic similarities to an extreme case of multisampling, \textit{complete multisampling}. However, multipooling is an optimization technique, whereas multisampling is a general technique for improving the capacity of CNNs. Multipooling suffers the same drawbacks as dilated convolutions in that neither technique performs any subsampling, blowing up the complexity of deep layers.

Zeiler and Fergus \cite{stochpool} use \textbf{stochastic pooling} as a regularization method for CNNs. Max-pooling and average-pooling is replaced with a stochastic pooling method that randomly samples an element from the pooling region according to a distribution given by the activities within the pooling region. Graham \cite{fmp} uses \textbf{fractional max-pooling} to randomly specify non-integer ratios between the spatial dimension sizes of the input and the output to pooling layers. Zhai et al. \cite{s3pool} use \textbf{S3Pool} which employs a deterministic pooling method followed by a stochastic downsampling method and is observed to have regularizing affects. 

Methods such as stochastic pooling, fractional max-pooling, and S3Pool focus on regularizing CNNs by implicitly increasing the size of the dataset through stochastic pooling methods. Multisampling also has strong regularizing effects during training, but differs fundamentally in that it addresses a different problem - the low spatial resolution of downsampled feature maps - and addresses it with a deterministic, algorithmic modification to explicitly increase the spatial resolution of feature maps. While fractional max-pooling may seem to share similarities at a glance, it does not decouple the height and width of feature maps from its spatial resolution and suffers the same fundamental drawbacks of traditional subsampling: The amount of spatial resolution lost in a subsampling layer scales quadratically with stride length.

\section{Problem Description}

Periodically throughout a CNN, feature maps pass through subsampling layers such as pooling layers and strided convolutional layers. Subsampling layers scale down the \textit{spatial} dimension lengths of the feature map so that the global receptive field of neurons in proceeding layers is increased. The magnitude of the downscale is determined by the stride length of the subsampling layer. While the spatial lengths scale down \textit{linearly} with stride length, resolution scales down \textit{quadratically} in a 2D CNN. In general, the new spatial resolution \(r_t\) of a feature map after passing through a CNN layer is: 
\begin{equation}
r_{t} = \frac{r_{t-1}}{k^d}
\label{eq:one}
\end{equation}
where \(r_{t-1}\) is the resolution before the layer is applied, \(k\) is the stride length, and \(d\) is the dimensionality of the CNN. For example, a subsampling layer with a stride length of 2 in a 2D CNN reduces spatial resolution by $4\times$.

Our goal is to design a subsampling scheme where the spatial resolution of the output feature map scales better with stride length and dimensionality while preserving the benefits of traditional subsampling layers such as increasing receptive field and reducing computational costs. This would have a number of benefits, including a more informative forward pass producing higher-resolution feature maps, better gradient updates for deep layers during training, and streamlining CNN design by reducing the need for dilated convolutions.

\begin{figure}[t]
\begin{center}
  \includegraphics[width=0.90\textwidth]{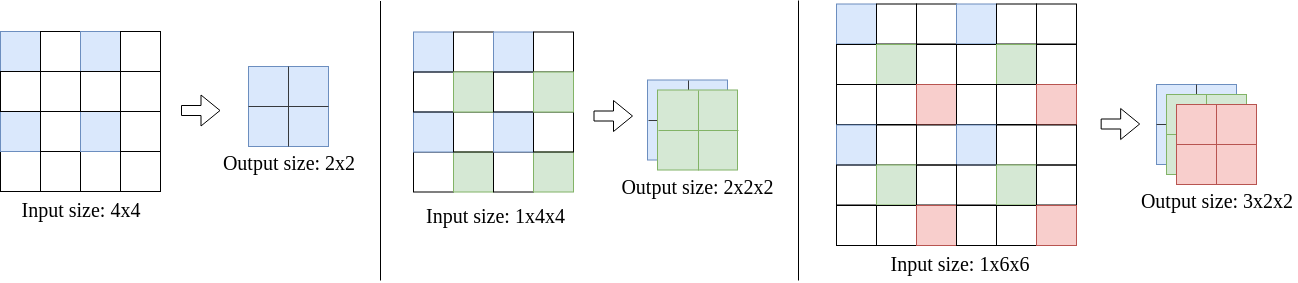}
  \caption{\textbf{Left}: The operation of a traditional CNN layer with stride $k = 2$. Blue highlighted elements represent locations kernels are applied, assume padding is used as necessary. \textbf{Middle}: The operation of a checkered subsampling layer. The new set of samples is stored as a separate submap. \textbf{Right}: One possible multisampling layer with $n = 3$ for layers with stride $k = 3$. Dimensions are read \#~submaps $\times$ height $\times$ width.}
\label{fig:one}
\end{center}
\end{figure}

\subsection{Solution: Multisampling}

One can imagine the operation of a traditional subsampling layer with a stride length of \(k\) in a 2D CNN as follows: First, the feature map is split into a grid of $k\times k$ \textbf{sampling windows}. Then, in each $k\times k$ sampling window, a pooling or convolutional operation is lined up with the top left element of the window (the blue highlighted elements in \cref{fig:one}) and the result of the operation becomes part of a new feature map. Our key insight is that one does not need to limit themselves to sampling only the top left corner of each sampling window. In a 2D CNN, we can choose up to \(k^2\) samples, multiplying the resolution of the output feature map by the number of samples taken \(n\). With this extension, which generalizes to higher dimensions, the new spatial resolution \(r_{t}\) of a feature map after passing through a CNN layer is: 
\vspace{-5pt}
\begin{equation}
r_{t} = \frac{n \cdot r_{t-1}}{k^d}
\label{eq:two}
\end{equation}
Our choice of where to sample from each sampling window is represented by a binary element-selector matrix termed the \textbf{sampler}. For example, in checkered subsampling we use a $2\times 2$ sampler that chooses the top-left and bottom-right element of each sampling window (\(n = 2\)). As traditional representations of feature maps do not have the capacity to store more than one sample from a sampling window, we extend feature maps with what we term a \textbf{submap dimension} and each sample is stored separately in its own feature submap across the submap dimension.

\begin{figure}[t]
	\centering
	\begin{subfigure}[t]{0.15\textwidth}
		\centering
		\includegraphics[width=2cm]{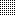}
        \caption*{$8\times8$}
	\end{subfigure}
	\begin{subfigure}[t]{0.15\textwidth}
		\centering
		\includegraphics[width=2cm]{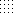}
        \caption*{$4\times4$}
	\end{subfigure}
 	\begin{subfigure}[t]{0.15\textwidth}
		\centering
		\includegraphics[width=2cm]{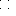}
        \caption*{$2\times2$}
	\end{subfigure}
    \rulesep
    \hspace{0.2cm}
   	\begin{subfigure}[t]{0.15\textwidth}
		\centering
		\includegraphics[width=2cm]{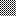}
		\caption*{$2\times 8 \times8$}
	\end{subfigure}
    	\begin{subfigure}[t]{0.15\textwidth}
		\centering
		\includegraphics[width=2cm]{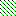}
		\caption*{$4\times 4 \times4$}
	\end{subfigure}
    	\begin{subfigure}[t]{0.15\textwidth}
		\centering
		\includegraphics[width=2cm]{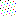}
		\caption*{$8\times 2 \times2$}
	\end{subfigure}
  \caption{\textbf{Left:} A $16\times16$ feature map is downsampled with stride = 2 once, twice, and three times with a traditional layer. \textbf{Right:} A $16\times16$ feature map is downsampled once, twice, and three times with a checkered layer, and feature maps are color coded so that elements belonging to the same submap share the same color. Each image is captioned with the dimension lengths of the resulting data structure. Notice \textit{every} row and column with respect to the original feature map is represented after each application of checkered subsampling. See supplementary materials for more illustrations.}
\label{fig:two}
\end{figure}

At each subsampling layer, multisampling is applied separately to each submap so that each submap is subsampled into \(n\) (number of samples taken by the sampler) new smaller submaps. Thus, the number of submaps is multiplied by $n$ times each time a multisampling layer that takes \(n\) samples is applied. All CNN layers such as convolutional, batch normalization \cite{batchnorm}, and dropout \cite{dropout}, layers are applied separately on each submap. After the final convolution, a CNN using multisampling layers will have generated many different submaps and several choices of post-processing are possible. In image classification, one can use a global 3D pooling layer (treating the submap dimension as a third spatial dimension) to generate a feature vector. If a 2D feature map is required, one may take the average across the submap dimension to generate a single submap which can be treated as a traditional feature map. Note that it is not necessary to process each submap independently of each other. One may use 3D convolutions to learn the best way to combine features across the submap dimension. 3D convolutions used in this way can learn deformed structures due to the semi-structured nature of the submap dimension. However, in most of our experiments, we process each submap independently.

We should be careful about our choice of samplers so that after many subsampling layers we obtain an efficient, well-distributed sampling of the input. One desirable property is for every row and column of the \textit{original image} to be represented by at least 1 sample and by the same number of samples. This is achieved if our sampler takes exactly one sample from each row and column of the \textit{sampling windows}. The minimum number of samples \(n\) from each sampling window required to accomplish this is exactly the stride length \(k\) and can be naively accomplished by sampling along a diagonal from opposite corners of a sampling window. In general, this can be accomplished by any n-rooks sampling \cite{nrooks} of the sampling window, which all take \(n=k\) samples. This value of \(n\) happens to have the very nice property of reducing the degree of the polynomial term in \cref{eq:two}.
\begin{equation}
r_{t} = \frac{k \cdot r_{t-1}}{k^d} = \frac{r_{t-1}}{k^{d-1}}
\end{equation}
In fact, in 2D CNNs, the exponent in the denominator is eliminated, resulting in spatial resolution scaling \textit{linearly}, rather than quadratically, with stride length \(k\). 
\begin{equation}
r_{t} = \frac{k \cdot r_{t-1}}{k^2} = \frac{r_{t-1}}{k}
\end{equation}
Thus, a n-rooks sampling of each sampling window is ideal as it provides the minimum number of samples needed to represent every row and column of each sampling window, and reduces the degree of the polynomial term in \cref{eq:two} so that resolution scales better with stride length (less information is lost). Finally, in order to ensure the samples are well-distributed, the same sampler should not be applied on each submap, even if the sampler satisfies the n-rooks property. This is because the final sampling will be biased by the choice of sampler and, after many subsampling steps, samples may aggregate in clumps or line-up in diagonals (see supplementary materials). One of two choices is possible: Randomly choose samplers that satisfy the n-rooks property each time a submap is subsampled in order to generate a random sampling of the input, or use a predetermined sequence of samplers to generate a low-discrepancy sampling of the input (we provide one such sequence using checkered subsampling samplers in the supplementary materials).

\subsection{Checkered subsampling}

By far, the most popular CNN architecture is a 2D CNN that uses subsampling layers with a stride length of 2. Therefore, we design \textbf{checkered subsampling} to replace the traditional subsampling layers of these models without affecting receptive fields. We call these converted models \textbf{checkered CNNs} (CCNNs). At each subsampling layer we sample the top-left and bottom-right element of each $2\times2$ sampling window (the blue and green elements in \cref{fig:one} respectively), satisfying the n-rooks property we desire in samplers. Each of the two samples is stored in a separate submap, so each application of checkered subsampling on a submap reduces it to 2 smaller submaps. Since we sample 2 of the 4 elements in a window, we keep 50\% of the input as opposed to 25\% with a traditional layer.

One may also use the \textit{complement} sampler where the top right and bottom left elements are sampled instead. By carefully applying one sampler to some submaps and the complement sampler to others, a regularly-spaced lattice sampling with respect to the original input can be obtained (see rightmost image of \cref{fig:two} and supplementary materials). Alternatively, by \textit{randomly} switching between the checkered sampler and its complement, a random sampling over the feature map can be obtained. Random switching during training may have regularizing properties by implicitly increasing the size of the dataset. However, in our experiments \textit{we do not use a random scheme}. Our goal is to show improvements during training come from the increased spatial capacity of feature maps, not from stochasticity introduced to training by a random subsampling scheme as in previous works \cite{fmp, stochpool,s3pool}. We use the simplest possible scheme in all of our image classification experiments, which is to apply the same sampler on every submap. Although using the same sampler on every submap biases the final samples to line-up in diagonals, we find this bias does not have a significant effect on accuracy in current architectures, which use small stride lengths and few subsampling layers.

\begin{figure}[t]
	\centering
	\begin{subfigure}[t]{0.15\textwidth}
		\centering
		\includegraphics[width=0.75cm]{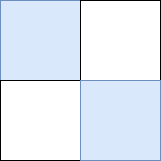}
	\end{subfigure}
	\begin{subfigure}[t]{0.15\textwidth}
		\centering
		\includegraphics[width=0.75cm]{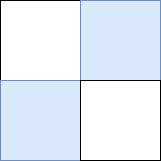}
	\end{subfigure}
  \caption{The two \textbf{samplers} used in checkered subsampling.}
\label{fig:three}
\end{figure}

\paragraph{CNNs versus CCNNs} By \cref{eq:one}, traditional subsampling layers with stride length \(k=2\), in 2D CNNs ($d=2$), reduce the resolution of their input by $4 \times$. Thus, an input image with resolution \(r_{input}\), after being processed by \(s\) subsampling layers, produces a feature map with resolution \(r_{trad}\):
\[r_{trad} = \frac{r_{input}}{4^s} = \frac{r_{input}}{2^s \times 2^s}\]
By \cref{eq:two}, in a 2D CCNN (which has \(n=2\) and \(k=2\)) each subsampling layer reduces the resolution of their input by $2 \times$. This means our advantage over traditional subsampling, in terms of the spatial capacity of resulting feature maps, grows \textit{exponentially }with each subsampling layer:
\[r_{checkered} = \frac{r_{input}}{2^s} = r_{trad} \times 2^s\]
Not only does this mean CCNNs produce drastically more informative feature maps in the later stages of the network than CNNs, but also exponentially increases the number of \textit{gradient updates} deep layers receive during training, as the number of gradient updates a layer receives is determined by the resolution of the input it gets. In our experiments we observe slightly faster convergence on CIFAR due to this.

The features generated by a CNN can be viewed as a subset those generated by a CCNN, thus CCNNs are theoretically guaranteed to offer superior representational capacity over CNNs with subsampling layers. To see this, imagine that an image has been processed by a CCNN, producing a feature map made up of $s$ submaps. If we throw away all but one submap and classify only on that one submap, we have reduced the capacity of our CCNN exactly to the capacity of a traditional CNN, and reintroducing any one additional submap pushes our capacity over that of a CNN. To see this visually, see the $2\times 2$ traditional feature map and $8\times2\times2$ checkered feature map in \cref{fig:two}. If we throw away every submap in the checkered feature map except for the black submap, we will be left with exactly the same samples produced by the traditional layers of a CNN.

\subsection{Relationship to traditional layers and dilation}

A $k\times k$ sampler that selects a single sample (\(n=1\)), the top left element, is exactly equivalent to a traditional 2D CNN layer with a stride length of \(k\). Thus, traditional CNN layers can be viewed as using an extreme version of multisampling where only the minimum number of samples needed to increase receptive field is taken. On the other hand, a $k\times k$ sampler that selects \textit{every} element to sample (\(n=k^2\) in a 2D CNN, or \(n=k^d\) in general), which we call \textbf{complete multisampling}, is functionally equivalent to not performing any subsampling and instead increasing the dilation of all proceeding layers by \(n\) times. This is because complete multisampling with a $k\times k$ sampler reduces the spatial lengths of all submaps by \(k\) times, and thus the receptive field of all proceeding neurons is increased by \(k\) times without performing any subsampling. The same effect is achieved by multiplying the dilation of the current layer and all proceeding layers by $n$ times. This is a common design choice in certain applications such as semantic segmentation which require fine-grained feature maps to reduce the blurriness of the generated output.  \cite{deeplabv3plus2018, semsegreview}. Thus, multisampling is a generalization of these techniques that enables finer control over how much information is lost at subsampling layers in-between these two extremes.

\section{Experiments and Discussion}
\label{Experiments}

We show that checkered subsampling drastically improves CNNs even in image classification, demonstrating for the first time (to the best of our knowledge) that coarse feature maps are bottlenecking the accuracy of these models. All experiments are performed on a single GTX 1080 Ti GPU. We reduce the memory requirements of large models during training with gradient checkpointing \cite{checkpointing}.

\subsection{Training current architectures as CCNNs}

We sample four popular architectures of different designs (VGG, ResNet, DenseNet, and Wide-ResNet) to train on CIFAR10 and CIFAR100. We write a conversion utility that takes 2D neural network layers as input (including convolutional, pooling, batch normalization, and dropout layers) and converts them into CCNN layers that can handle and process submaps. \textbf{Note that no parameters are added in this process.} Layers with a stride length of 2 are modified to use checkered subsampling. After the final convolution, all submaps are averaged into a single submap / feature map which is fed into an unmodified classifier. We train our models before and after applying our conversion utility. Our CIFAR models use 2 (DenseNet, Wide-ResNet), 3 (ResNet) or 5 (VGG) subsampling layers, so our CCNNs increase the amount of information in the final feature maps by \(2^2\), \(2^3\), or \(2^5\) times.

CIFAR10 consists of 50,000 training images and 10,000 test images from 10 classes. CIFAR100 consists of 50,000 training images and 10,000 test images from 100 classes. Classes include common objects such as cat, dog, automobile, and airplane. DenseNet for CIFAR is obtained from the implementation of Pleiss et al. \cite{memoryefficientdensenet}. VGG and ResNet for CIFAR are obtained from \cite{cifar-pytorch}. Wide-ResNet is obtained from \cite{wide-resnet-pytorch}. For ResNet and VGG we increase the batch size to 128 and decrease number of epochs to 164 as in their original descriptions \cite{resnet, vgg} and use the training script of Pleiss et al. Otherwise all hyperparameters are left at default values - no hyperparameter tuning is performed. We train on all training images and report accuracy on test images. For data augmentations we use the standard scheme: We randomly apply horizontal flips and randomly shift horizontally or vertically by up to 4-pixels.

\begin{table}[t]
  \centering
  \caption{Test error on CIFAR after training as a CNN and as a CCNN. The asterisk (*) indicates \textit{without} data augmentations. Conversion to a CCNN significantly improves all models we test. We also run an experiment on Wide-Resnet where we perform convolutions across two submaps.}
  \begin{tabular}{l|cc|cc|cc}
    \toprule
    Architecture     	   	 	& \multicolumn{2}{c|}{C10*} & \multicolumn{2}{c|}{C10} & \multicolumn{2}{c}{C100} \\[-0.8ex]
    & \tiny CNN & \tiny CCNN & \tiny CNN & \tiny CCNN & \tiny CNN & \tiny CCNN \\[-0.5ex]
    \midrule
    DenseNet-BC-40     			& 9.13    			& \textbf{7.77}	 & 6.73    			& \textbf{6.49}			& 29.32    			& \textbf{28.55}						\\
    
    DenseNet-BC-121     		& 6.56      			& \textbf{5.37}	 & 4.19      			& \textbf{3.95}		   		& 20.32      			& \textbf{19.97}	   			\\
    
    ResNet-18     				& 12.81		 		& \textbf{9.90} & 5.49		 		& \textbf{4.90} 	   				& 25.70	 			& \textbf{24.95}					\\
    
    ResNet-50     				& 12.11		 		& \textbf{10.68} & 5.31		 		&\textbf{5.17}	   				& 24.75	 			& \textbf{22.21}		    				\\
    VGG-11-BN					& 14.62				& \textbf{11.57} & 8.23				& \textbf{7.47}						& 29.93				& \textbf{28.97}					\\
    Wide-Resnet-28x10	 &   &   & 3.80		& \textbf{3.60} & 18.89  & \textbf{18.74} \\
    \hspace{0.25cm} \small{+ $2\times$3$\times$3 convolutions} &   &   &   & \textbf{3.51} &   &   \\
    \bottomrule
  \end{tabular}
  \label{table:one}
\end{table}

We find checkered subsampling gives a significant performance boost to \textit{every} model we train (\cref{table:one}). Interestingly, we observe that a ResNet-18 CCNN \textit{outperforms} the deeper ResNet-50 CNN \textit{and} CCNN on CIFAR10, although the ResNet-50 CCNN receives a significant performance boost over the ResNet-18 CCNN on CIFAR100. We also experiment with applying 3D convolutions across the submap dimension. In Wide-ResNet, we replace all $3\times3$ convolutions after the second subsampling layer with $2\times3\times3$ convolutions. We observe that a 28 layer Wide-ResNet CCNN extended with 3D convolutions is competitive with a 164 layer PyramidNet \cite{pyramidnet} on CIFAR10.

\begin{wrapfigure}{r}{0.42\textwidth}
  \vspace{-30pt}
  \begin{center}
    \includegraphics[width=0.40\textwidth]{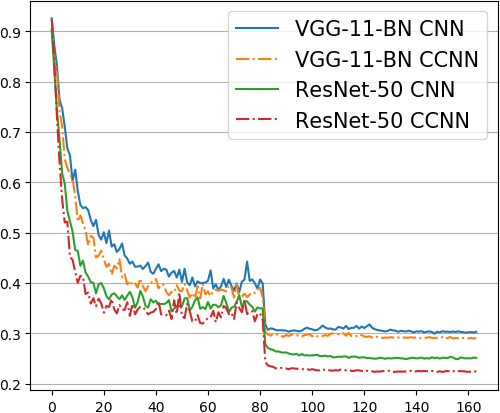}
  \end{center}
  \caption{Test curves on CIFAR100.}
  \vspace{-15pt}
\label{fig:four}
\end{wrapfigure}

We also notice all models show steeper test curves as CCNNs than as CNNs (\cref{fig:four}), with the effect more pronounced on the CIFAR100 dataset. One reason for this may be that CCNNs provide much more gradient updates to deep layers than CNNs. Each subsampling layer in a CNN reduces the number of samples (and thus the number of gradients) proceeding layers will receive by $4\times$, whereas checkered subsampling layers reduce the number of gradients by only $2\times$.

\paragraph{Multisampling versus data augmentations} We observe that the benefits of checkered subsampling is amplified when data augmentations are disabled. DenseNet, which was observed in \cite{densenet} to train exceptionally well without data augmentations, receives a further 1.36\% and 1.19\% absolute performance boost on CIFAR10. 

These results glean new insight into the nature of data augmentations. We argue data augmentations allow deep CNN layers to \textit{see} information that they would not have otherwise seen by giving feature detectors a chance to line-up with all image features over many epochs of training. The left 3 images in \cref{fig:two} illustrates how after a few traditional subsampling layers, deep convolutions are very limited in where they are allowed to be applied with respect to the original image. Convolutions work best when they are centered on the features they are trying to detect, so it is necessary to feed the same image many times under many different augmentations before deep feature detectors receive a good sampling of the input. Multisampling reduces the need for repeated exposures under different augmentations by explicitly extracting that unseen information within a single forward pass. This is why stochastic subsampling methods \cite{fmp, stochpool, s3pool} are also observed to have strong regularizing effects in the absence of data augmentations: They are able to sample parts of the feature map that would not have otherwise been considered for training over many epochs.

\subsection{Multisampling pretrained models without any training}

We convert publicly available models pretrained on ImageNet into CCNNs by applying our CCNN conversion utility on each model. We transfer over the parameters of the original CNN into its CCNN counterpart without any modifications. This can be done because checkered subsampling is an algorithmic change in how striding is performed and how feature maps are stored and is compatible with the learned kernels of a traditional CNN. Our converted ImageNet models can be viewed as extracting an ensemble of \(2^n\) feature submaps, where \(n\) is the number of subsampling layers (\(n=5\) in most ImageNet models, \(n=4\) in SqueezeNet). Unlike traditional ensembles, this ensemble is produced from a single image by a single model, requires only a single forward pass, and can be produced by any arbitrary CNN with subsampling layers. ImageNet models tend to follow a common pattern: A series of convolutional layers followed by a fully-connected classifier. After the final convolution, the multisampled feature submaps (i.e., the ensemble of feature maps) must be converted into a form that the final pretrained classifier can handle. We simply produce an average feature map by taking the mean across the submap dimension and feed the averaged feature map into the classifier.

We use the ILSVRC2012 validation dataset as in \cite{densenet}, which consists of 50,000 images sampled from ImageNet with 1,000 different classes, to test the performance of pretrained models before and after the conversion. \textbf{No training, fine-tuning, or modification of model parameters is performed.} All pretrained models except for FB-ResNet are obtained from torchvision \cite{torchvision}. FB-ResNet is obtained from \cite{fbres}.

\begin{table*}[t]
   \caption{Top-1 single-crop errors of publicly available pretrained ImageNet models before and after transferring parameters to CCNN layers \textbf{without any training or fine-tuning}. Certain types of models are hurt by this transformation because they don't generalize well to an alternative spatial resolution of the final feature maps, as explained below.}
  \begin{tabular}{l|cc}
    \toprule
    Architecture     		& CNN  	& CCNN\\
    \midrule
    AlexNet 		& \textbf{43.48}    & 43.97	\\
    \midrule
    DenseNet-121		& 25.57    & \textbf{25.55}\\
    DenseNet-161		& 22.86    & \textbf{22.77}\\
    DenseNet-169		& 24.40    & \textbf{24.00}\\
    \midrule
    ResNet-101 			& 22.63    & \textbf{22.47}\\
    ResNet-152			& 21.87    & \textbf{21.57}\\
    FB-ResNet-152 		& 22.61   & \textbf{22.38}\\
    \bottomrule
  \end{tabular}
\hspace{0.5cm}
\begin{tabular}{l|cc}
    \toprule
    Architecture     		& CNN  	& CCNN\\
    \midrule
    ResNet-18 			& \textbf{30.24}    & 30.66\\
    ResNet-34 			& \textbf{26.69}    & 26.85\\
    ResNet-50 			& \textbf{23.87}    & 23.90 \\
    \midrule
    SqueezeNet-1.0	 	& 41.91    & \textbf{41.64}\\ 
    SqueezeNet-1.1 		& 41.82    & \textbf{41.28}\\ 
    \midrule
    VGG-11		 		& \textbf{30.98}    & 31.39 \\
    VGG-19 				& \textbf{27.62}  	& 28.18 \\
    \bottomrule
  \end{tabular}
  \label{table:two}
\end{table*}

We find converting pretrained ImageNet models to CCNNs, without any training or tuning, improves the top-1 performance of certain classes of models. Deeper ResNet models benefit more than shallow models. This pattern is clear going down the results of the ResNet models, with ResNet-18 showing the worst impact (0.42\% worse performance) and ResNet-152 / FB-ResNet-152 showing the best impact (0.30\% and 0.23\% performance boost respectively). We also observe that smaller models (in terms of parameter count) benefit more than larger models when the depth of the network is similar. For example, both versions of SqueezeNet receive a considerable performance boost, but the lighter SqueezeNet1.1 receives a significantly larger boost of 0.54\% compared to the boost of 0.27\% for SqueezeNet1.0. The pretrained DenseNet models provided by torchvision use different growth rates. DenseNet-161 uses a growth rate \(k=40\), while DenseNet-169 uses a smaller growth rate \(k=24\). The result is that DenseNet-169 uses significantly less parameters.

Pretrained AlexNet and VGG models are hurt by the conversion to checkered subsampling in all our experiments. We believe these models are too fine-tuned to the expected spatial distribution of features to benefit from this technique because they do not use a global pooling layer before their final classifier. In particular, the classification layers in AlexNet and VGG expect the feature maps to have been computed with padding at certain edges, and our technique disturbs the location of padding.

We are not sure why shallow ResNet model are hurt by the conversion while deep ResNet models benefit from it. Nor why the parameter-count of the model might effect its robustness to this conversion. Possibly, large models of a certain class may have learned to "overfit" to their original downsampling scheme.

\paragraph{Checkered subsampling versus dilation} We describe an alternative strategy for producing denser feature maps from pretrained models by using dilated layers. One can set the stride length of all subsampling layers from \(k\) to 1, and instead increase the dilation of all proceeding layers by \(k\) times, taking care not to lose information at edges by increasing padding correspondingly. Similarly, one may perform complete multisampling at each subsampling layer, which has the same effect as the method using dilation. We find that due to the lack of any subsampling, these methods are extremely computationally expensive. Furthermore, despite producing denser feature maps, dilated layers and complete multisampling \textbf{do not} offer a significant accuracy boost over checkered subsampling in this task due to diminishing returns.

\begin{table}[h]
  \tabcolsep=0.19cm
  \centering
  \caption{Inference time, memory consumption, and error of pretrained ImageNet models before and after conversion to a checkered CNN or a dilated CNN (with batch size of 4 on a GTX 1080 Ti).}
  \begin{tabular}{l|ccc|ccc|ccc}
    \toprule
    Type & \multicolumn{3}{c|}{SqueezeNet-1.1} & \multicolumn{3}{c|}{ResNet-152} & \multicolumn{3}{c}{DenseNet-169}\\
    \midrule
    Original 				&  0.007 s & 0.6 GB	& 41.82		& 0.02 s & 0.9 GB & 21.87 		& 0.02 s  & 0.8 GB & 24.40\\
    Dilated 				&  0.15 s  & 3.2 GB & 41.31		& 3.60 s & 10.1 GB & 21.60		& 1.67 s  & 10.6 GB & 23.98\\
    \textbf{Checkered} 		&  0.02 s  & 0.7 GB & 41.28		& 0.25 s & 1.2 GB & 21.57	 	& 0.11 s  & 2.2 GB & 24.00\\
    \bottomrule
  \end{tabular}
  \label{table:three}
\end{table}

\section{Conclusion}
We show that there is a significant amount of spatial information that current subsampling layers fail to utilize and show that we can use a simple and efficient implementation of multisampling, checkered subsampling, to extract that information to improve the learning and accuracy of CNNs. We significantly improve the accuracy of every image classification model we train, demonstrating that the limited spatial capacity of coarse feature maps is bottlenecking current architectures. We improve the accuracy of certain classes of pretrained ImageNet models without any training or fine-tuning by simply applying multisampling. We observe that the benefit of checkered subsampling is amplified when no data augmentations are used, leading to our argument that the effectiveness of data augmentations is in major part due to the extra spatial information they extract from images over many epochs. We believe multisampling will find further use in applications where fine-grained information is important, such as semantic segmentation and in generative models, where multisampling-based techniques may generate finer outputs and serve as an efficient alternative to dilation. Our code is public at \url{https://github.com/ShayanPersonal/checkered-cnn}.

\section*{Supplementary Materials}

\paragraph{Implementation} We give a brief description of implementing checkered subsampling here and our code is also available on Github at \url{https://github.com/ShayanPersonal/checkered-cnn}. The operation of each strided layer needs to be modified to apply a checkered sampler or its complement when stride = 2. For the standard checkered sampler this can be achieved by applying the operation as normal to the input feature map \(x\), shifting the input feature map \(x\) by one element in both spatial dimensions, and applying the operation again. This generates a total of two output feature maps (in this case, feature \textit{submaps}) which are then concatenated together along the submap dimension to create the output feature map \(y\).
\begin{equation}
y = Concatenate(Conv(x), Conv(ShiftRight(ShiftDown(x))))
\end{equation}
Similarly, the function for applying the complement sampler is: 
\begin{equation}
y = Concatenate(Conv(ShiftRight(x)), Conv(ShiftDown(x)))
\end{equation}
CNN operations should be applied independently on all submaps. A naive way to achieve this is to apply 2D layers separately on each submap, but in practice this is inefficient as each layer needs to be re-executed for every submap. A more efficient implementation is to replace all 2D layers with their equivalent 3D counterparts and add a submap dimension on the input to the CNN in place of where 3D layers expect the depth dimension to be. That is, all $m\times m$ kernels should be replaced with $1 \times m\times m$ kernels and similarly all stride lengths should be modified from $k\times k$ to $1\times k\times k$. All submaps will then be processed in a single pass through a 3D layer, rather than many passes through a 2D layer. This formulation also enables the use of 3D convolutions if desired.

\paragraph{Complexity} Consider a ResNet-style architecture where we start off with a base number of feature channels at the earliest layer (e.g., 32), the number of feature channels is increased by $2\times$ after each subsampling layer, and each subsampling layer increases receptive field by $2\times$. Suppose we design our network using either traditional subsampling, checkered subsampling, or don't use subsampling and instead increase dilation by $2\times$. We can compute the effect each method has on the complexity of proceeding layers and show checkered subsampling falls in the middle-ground between traditional layers and dilated layers.

\begin{table}[h]
  \tabcolsep=0.19cm
  \centering
  \caption{Complexity of a layer in a ResNet-style model (where the number of channels is increased by $2\times$ at each subsampling step) in terms of the number of subsampling layers, \(s\), preceding the layer. }
  \begin{tabular}{lcc}
    \toprule
    Subsampling layer type & Memory complexity & Compute complexity \\
    \midrule
    Traditional 			&  $0.5^s$ & $1$			\\
    \textbf{Checkered} 		&  $1$  & $2^s$ 	\\
    Dilated 				&  $2^s$  & $4^s$	\\
    \bottomrule
  \end{tabular}
  \label{table:blah}
\end{table}

\begin{table}[h]
  \tabcolsep=0.19cm
  \centering
  \caption{Complexity of a layer in an architecture where the number of channels is kept constant after subsampling, in terms of the number of subsampling layers preceding the layer, \(s\).}
  \begin{tabular}{lcc}
    \toprule
    Subsampling layer type & Memory complexity & Compute complexity \\
    \midrule
    Traditional 			&  $0.25^s$ & $0.25^s$			\\
    \textbf{Checkered} 		&  $0.5^s$  & $0.5^s$ 	\\
    Dilated 				&  1  & 1	\\
    \bottomrule
  \end{tabular}
  \label{table:blah}
\end{table}

In practice, on CIFAR we observed about $1.1\times$ to  $2\times$ increased memory usage and about $2\times$ to $6\times$ increased training time converting current architectures. \cref{table:three} shows that inference time on ImageNet models increases anywhere from around $3\times$ to $13\times$. Note these results are obtained with our unoptimized implementation using high-level Pytorch operations.

We suspect that one reason so many channels are required in the late stages of current architectures is to "remember" information that is deleted by subsampling. This would explain why DenseNet performs well with so few parameters compared to other architectures - its skip connections preserve fine-grained details that would otherwise be lost, so it does not need so many channels at every step to remember those details. Architectures built on top of checkered subsampling layers may be able to reduce the number of channels in deep layers of their architecture and still obtain state-of-the-art results. In order to maintain a constant compute complexity with checkered subsampling, the number of channels should be multiplied by $\sqrt{2}$, or \textasciitilde1.41, after each subsampling layer.

\begin{table}[h]
  \tabcolsep=0.19cm
  \centering
  \caption{To maintain constant compute costs with checkered subsampling, the number of channels should be multiplied by $\sqrt{2}$ after each subsampling layer to achieve the following complexity:}
  \begin{tabular}{lcc}
    \toprule
    Subsampling layer type & Memory complexity & Compute complexity \\
    \midrule
    \textbf{Checkered} 		&  $\frac{1}{2^{s/2}}$  & $1$ 	\\
    \bottomrule
  \end{tabular}
  \label{table:blah}
\end{table}

To test our hypothesis, we modify ResNet to use our $\sqrt{2}$ scaling rule. At each subsampling step, the number of channels is increased by $\sqrt{2}$ (64, 91, 128, 181) rather than by 2 as in the original ResNet (64, 128, 256, 512). For the bottleneck layer of ResNet-50, we reduce the expansion factor from four to two. We find that our tiny ResNet models, trained as a CCNN, are competitive with or better than their full-sized CNN counterparts on CIFAR100 with augmentations.

\begin{table}[h]
  \centering
  \caption{Our tiny ResNet CCNNs are competitive with / better than their full-sized CNN counterparts.}
  \begin{tabular}{l|c|cc}
    \toprule
    Architecture     	   	 	& Parameter count  & \multicolumn{2}{c}{C100 Error} \\[-0.8ex]
    & & \tiny CNN & \tiny CCNN \\[-0.5ex]
    \midrule
    ResNet-18     			& 11.2M					& 25.70	 			& \textbf{24.95}			\\
    ResNet-18-tiny     		& 2.1M  				& 26.74	 			& \textbf{25.68}	\\
    ResNet-50     			& 23.5M	   			& 24.75	 			& \textbf{22.21}	\\
    ResNet-50-tiny     		& 3.3M				& 26.12	 			& \textbf{24.17}				\\
    \bottomrule
  \end{tabular}
  \label{table:9}
\end{table}

Next, we create a toy neural network to train on MNIST \cite{mnist} with 5 convolutional layers of 32, 32, 45, 45, 64 channels followed by a linear classifier. The 3rd layer performs subsampling with a stride length of 2. As a CCNN the layer performs checkered subsampling and outputs 2 submaps. Each layer is followed by batch normalization \cite{batchnorm}. Dropout \cite{dropout} with a rate of 0.2 is applied before the linear classifier. We train our network both as a CNN and as a CCNN for 100 epochs with SGD with Nesterov momentum factor of 0.9 and batch size of 16. We report the best single-run accuracy observed after training without data augmentations, with shift-only data augmentations of up to 2 pixels as in \cite{capsules}, and with both shift augmentations and rotational augmentations of up to 15 degrees. As a CCNN we also test $2\times3\times3$ at the 5th layer which learns to combine the two submaps into one.

We observe that checkered subsampling improves accuracy in all cases. For comparison, we include the results of Sabour et al. \cite{capsules} which claims to be state-of-the-art on MNIST. Our CCNN outperforms the CNN baseline used in \cite{capsules}, which has $553\times$ more parameters, under the same augmentation scheme. Our extended CCNN is competitive with a capsule network unaided by a reconstruction network, which has $73\times$ more parameters. With 15 degree rotational augmentations, our CCNN is competitive with a capsule net with its reconstruction network, which has $88\times$ more parameters. We train our best CCNN 5 times and estimate the mean score and standard deviation. The errors we observed in 5 trials ordered by accuracy are 0.23, 0.23, 0.25, 0.27 and 0.27. To the best of our knowledge in 2018, this is among the best reported results on MNIST for a single neural network without ensembling.

\begin{table}[h]
  \centering
  \caption{We create toy CNNs to test on MNIST and report their errors. We include state-of-the-art results from \cite{capsules} for comparison.}
  \begin{tabular}{l|c|ccc}
  	\toprule
    Architecture     	   	 	& Parameters  & Error (no aug) & Error (shift aug) & Error (shift+rot) \\
    \midrule
    CNN baseline of \cite{capsules} & 35.4M & - & 0.39 & - \\
    CapsNet w/o reconstruct & 6.8M & - & 0.34 & - \\
    CapsNet w/ reconstruct & 8.2M & - & $0.25_{\pm 0.005}$ & - \\
    \midrule
    Tiny CNN				& 67,913 & 0.44 & 0.42 & \textbf{0.30}	\\
    Tiny CCNN				& 67,913 & 0.39 & 0.38 & \textbf{0.28} 	\\
    Tiny CCNN w/ $2\times3\times3$		& 93,833 & 0.39 & 0.35 & {\boldmath $0.25_{\pm 0.02}$} 	\\
  \bottomrule
  \end{tabular}
  \label{table:9}
\end{table}

\paragraph{Low-discrepancy sampling and other patterns}
We discuss instances of checkered subsampling and its implementation. Multisampling is not limited to layers with stride = 2. We also depict an algorithm for layers with stride = 3 that preserves 33\% of the input map resolution at each subsampling step (in contrast to 11\% without multisampling).

First we discuss how to generate a low-discrepancy lattice sampling of the input using checkered subsampling. Consider \cref{fig:latticesmall}:

\begin{figure}[h]
\begin{center}
  \includegraphics[width=3.5cm]{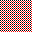}
  \vspace{0.5cm}
  \hspace{0.5cm}
  \includegraphics[width=3.5cm]{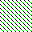}
    \hspace{0.5cm}
  \includegraphics[width=3.5cm]{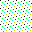}
    \hspace{0.5cm}
  \includegraphics[width=3.5cm]{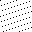}
    \hspace{0.5cm}
  \includegraphics[width=3.5cm]{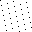}
   \caption{A 32x32 image undergoes checkered subsampling 1, 2, 3 (top row), 4, and 5 (bottom row) times with our low-discrepancy lattice sequence. In the first 3 images, features belonging to the same submap are colored identically to help with the intuition.}
  \label{fig:latticesmall}
 \end{center}
\end{figure}

\begin{figure}[h]
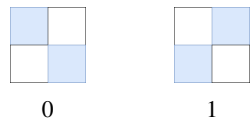

	\centering
	\begin{subfigure}[t]{0.15\textwidth}
		\centering
		\includegraphics[width=1cm]{check_tri.png}
        \caption*{0}
	\end{subfigure}
	\begin{subfigure}[t]{0.15\textwidth}
		\centering
		\includegraphics[width=1cm]{check_tri__1_.png}
        \caption*{1}
	\end{subfigure}
  \caption{The two samplers used in checkered subsampling can be identified by a binary value.}
\label{fig:yo}
\end{figure}

In order to generate these samplings, a checkered sampler (which samples the top-left and bottom-right sample in a sampling window) had to be applied on certain submaps and the complement checkered sampler (samples the top-right and bottom-left sample in a sampling window) had to be applied on others. Suppose a 0 represents a checkered sampler and a 1 represents the complement sampler. The above images were generated with the following sequence:

0 \\
0, 0 \\
0, 1, 0, 1 \\
0, 1, 1, 0, 0, 0, 1, 1\\
0, 0, 1, 0, 1, 0, 0, 1, 0, 1, 0, 0, 1, 0, 1, 0\\

Here's how to read this sequence. The \textbf{first line} says apply a checkered sampler onto the original input (not depicted) to obtain the black and red submaps in the top left image. The \textbf{second line} describes how to process the top left image to obtain the sampling in the middle image of the top row, and says apply a checkered sampler to the black submap to obtain the black and green submaps, and a checkered sampler to the red submap to obtain the red and blue submaps. So far we have applied the same sampler to every submap.

The \textbf{third line} of the sequence describes how to process the middle image in the top row to obtain the top right image. It says apply a checkered sampler to the black submap to obtain the black and cyan submaps, a \textit{complement} checkered sampler to the red submap to obtain the red and purple submaps, a checkered sampler to the green submap to obtain the green and yellow submaps, and a \textit{complement} checkered sampler to the blue submap to obtain the blue and grey submaps.

The fourth line of the sequence is then used to process the top right sampling into the bottom left sampling, and the fifth line is used to process the bottom left sampling into the bottom right sampling.

In general, the length of each line is the number of submaps represented before applying the samplers listed on the line. Each value indicates which type of sampler to use on that submap at the next subsampling step. The first value corresponds to the submap containing the topmost row of the image and each subsequent value corresponds to the submap containing the next row of the image going down. This works because by our construction of multisampling, every row (and column) is represented by exactly one submap.

\begin{figure}[h]
\begin{center}
  \includegraphics[width=6.5cm]{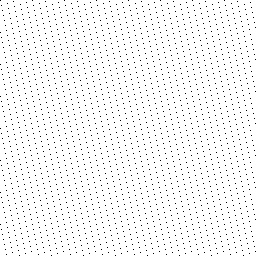}
  \vspace{0.2cm}
  \hspace{0.2cm}
  \includegraphics[width=6.5cm]{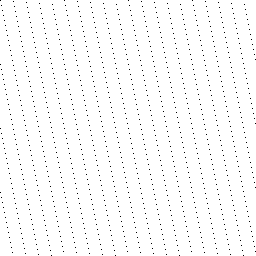}
    \hspace{0.2cm}
  \includegraphics[width=6.5cm]{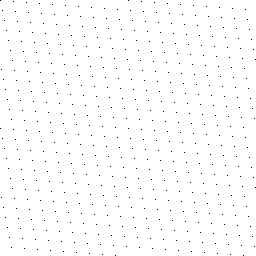}
    \hspace{0.2cm}
  \includegraphics[width=6.5cm]{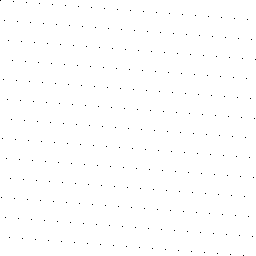}
  \caption{A 256x256 image undergoes our low-discrepancy lattice sequence through 5, 6 (top row), 7, and 8 (bottom row) checkered subsampling layers.}
  \label{fig:256even}
 \end{center}
\end{figure}

We continue the previous low-discrepancy lattice sequence to 10 subsampling steps. See \cref{fig:256even} for a higher-resolution depiction of our lattice sequence up to 8 subsampling steps. Those familiar with quasi-Monte Carlo methods may be reminded of tables of parameters for the construction of good lattice points found in the literature on integration lattice techniques (see section 6 of \cite{quasi}, \textit{Quasi-Monte Carlo Sampling} by Owen.)

\textbf{(1)} 0\\
\textbf{(2)} 00\\
\textbf{(3)} 0101\\
\textbf{(4)} 01100011\\
\textbf{(5)} 0010100101001010\\
\textbf{(6)} 00011000110001100011000110001100\\

\textbf{(7)} \seqsplit{0000011111000001111100000111110000011111000001111100000111110000}

\textbf{(8)} \seqsplit{00000000001111111111000000000011111111110000000000111111111100000000011111111110000000000111111111100000000001111111111000000000}

\textbf{(9)} \seqsplit{0000000000000000000011111111111111111111000000000000000000001111111111111111111000000000000000000001111111111111111111100000000000000000001111111111111111111100000000000000000000111111111111111111100000000000000000000111111111111111111110000000000000000000}

\textbf{(10)} \seqsplit{010101010101010101010101010101010101010101010101010101010101010101010101010101010101010101010101010101010101010101010101010101010101010101010101010101010101010101010101010101010101010101010101010101010101010101010101010101010101010101010101010101010101010101010101010101010101010101010101010101010101010101010101010101010101010101010101010101010101010101010101010101010101010101010101010101010101010101010101}

\clearpage

\begin{figure}[h]
\begin{center}
  \includegraphics[width=11cm]{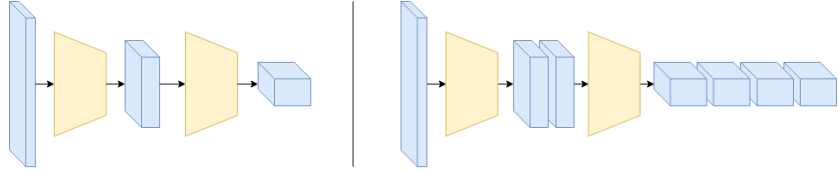}
  \caption{A high-level flow of a CNN (left) and CCNN (right) is depicted. Blue rectangular prisms represent feature maps and yellow trapezoids represent groups of convolutional layers with a subsampling layer of stride = 2 present. The CCNN downscales the feature maps to the same spatial dimensions as the CNN but preserves more spatial information through the use of feature submaps. Note that while the number of submaps scales exponentially with the number of subsampling steps, the height and width of feature maps both decrease exponentially, leading to an overall exponential decrease in resolution akin to traditional subsampling.} 
 \end{center}
\end{figure}

\begin{figure}[h]
If one applies the same sampler repeatedly to all submaps, the final sampling will be biased so that samples form into clumps or diagonals. \cref{fig:naive} shows what samplings look like if we only use checkered subsampling without its complement (i.e., generate rows of 0's only).
\vspace{0.2cm}
\begin{center}
  \includegraphics[width=3.5cm]{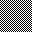}
  \vspace{0.5cm}
  \hspace{0.5cm}
  \includegraphics[width=3.5cm]{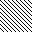}
    \hspace{0.5cm}
  \includegraphics[width=3.5cm]{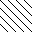}
    \hspace{0.5cm}
  \includegraphics[width=3.5cm]{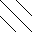}
   \caption{A 32x32 image undergoes checkered subsampling 1, 2, 3 (top row), and 4 (bottom row) times with the same checkered sampler applied to every submap at every step. When checkered subsampling is naively applied this way features begin to line-up in diagonals. This method still offers superior resolution over traditional subsampling layers (which would be left with only 4 samples after 4 subsampling layers) and works very well in our experiments, but may not be ideal in applications that need to generate fine images from feature maps such as in semantic segmentation.}
  \label{fig:naive}
 \end{center}
\end{figure}

\clearpage

\begin{figure}[h]
\vspace{0.2cm}
\begin{center}
  \includegraphics[width=3.5cm]{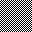}
  \hspace{0.5cm}
  \vspace{0.5cm}
  \includegraphics[width=3.5cm]{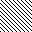}
    \hspace{0.5cm}
  \includegraphics[width=3.5cm]{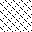}
    \hspace{0.5cm}
  \includegraphics[width=3.5cm]{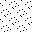}
     \hspace{0.5cm}
  \includegraphics[width=3.5cm]{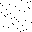}
   \caption{A 32x32 image undergoes checkered subsampling under a random sequence.}
 \label{fig:rand1}
\end{center}
\end{figure}

\begin{figure}[h]
\begin{center}
  \includegraphics[width=3.5cm]{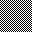}
  \hspace{0.5cm}
  \vspace{0.5cm}
  \includegraphics[width=3.5cm]{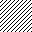}
    \hspace{0.5cm}
  \includegraphics[width=3.5cm]{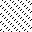}
    \hspace{0.5cm}
  \includegraphics[width=3.5cm]{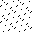}
     \hspace{0.5cm}
  \includegraphics[width=3.5cm]{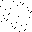}
   \caption{A 32x32 image undergoes random checkered subsampling using a different seed.}
 \label{fig:rand2}
\end{center}
\end{figure}

\begin{figure}[h]
\begin{center}
  \includegraphics[width=6.5cm]{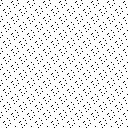}
  \vspace{0.2cm}
  \hspace{0.2cm}
  \includegraphics[width=6.5cm]{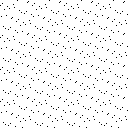}
    \hspace{0.2cm}
  \includegraphics[width=6.5cm]{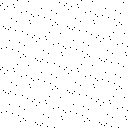}
    \hspace{0.2cm}
  \includegraphics[width=6.5cm]{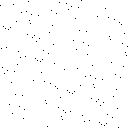}
  \caption{A 128x128 image is subsampled through 4, 5 (top row), 5, and 6 (bottom row) checkered subsampling layers with a randomly generated sequence.}
  \label{fig:randbig}
 \end{center}
\end{figure}

Alternatively, we can randomly generate sequences of 0's and 1's to randomly apply one of the two samplers on each submap. \cref{fig:rand1}, \cref{fig:rand2}, and \cref{fig:randbig} show how the process of subsampling looks when samplers are randomly applied. Due to the existence of regularly spaced lattice sequences, it is possible to engineer your own sequences to be close to regularly-spaced.

\clearpage

\clearpage

\clearpage

\begin{figure}[h]
Some sequences generate interesting patterns over the original image. We depict some patterns observed after 4-5 random steps of checkered subsampling.
\begin{center}
  \includegraphics[width=6.5cm]{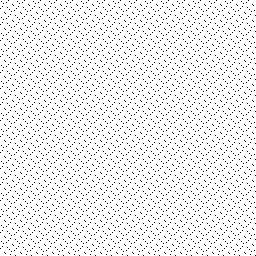}
  \vspace{0.2cm}
  \hspace{0.2cm}
  \includegraphics[width=6.5cm]{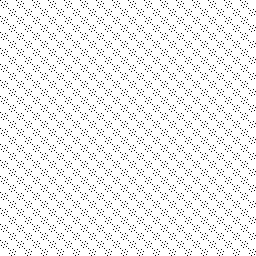}
  \hspace{0.2cm}
  \includegraphics[width=6.5cm]{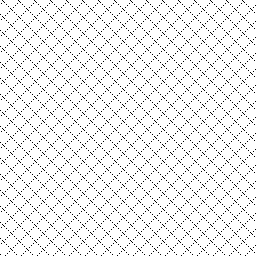}
  \vspace{0.2cm}
  \hspace{0.2cm}
  \includegraphics[width=6.5cm]{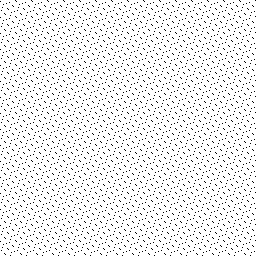}
  \hspace{0.2cm}
  \includegraphics[width=6.5cm]{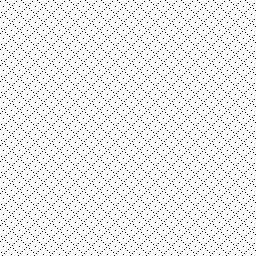}
  \vspace{0.2cm}
  \hspace{0.2cm}
  \includegraphics[width=6.5cm]{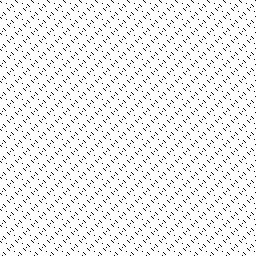}
  \hspace{0.2cm}
 \end{center}
\end{figure}

\begin{figure}[h]
\begin{center}
  \includegraphics[width=6.5cm]{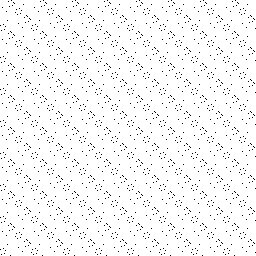}
  \vspace{0.2cm}
  \hspace{0.2cm}
  \includegraphics[width=6.5cm]{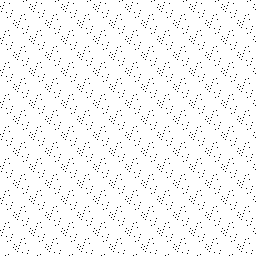}
  \hspace{0.2cm}
  \includegraphics[width=6.5cm]{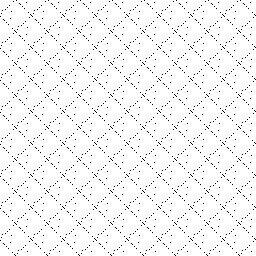}
  \hspace{0.2cm}
 \end{center}
\end{figure}

\clearpage

\begin{figure}[h]
\begin{center}
  \includegraphics[width=8cm]{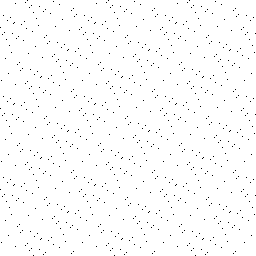}
  \hspace{0.2cm}
  \caption{Multisampling generalizes to larger stride lengths. This pattern was generated by randomly applying one of three $3\times 3$ samplers with \(n=3\).}
  \label{fig:tri}
 \end{center}
\end{figure}

\begin{figure}[h]
	\centering
	\begin{subfigure}[t]{0.20\textwidth}
		\centering
		\includegraphics[width=2cm]{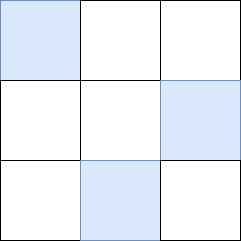}
        \caption*{0}
	\end{subfigure}
	\begin{subfigure}[t]{0.20\textwidth}
		\centering
		\includegraphics[width=2cm]{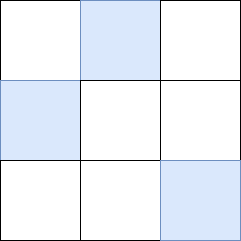}
        \caption*{1}
	\end{subfigure}
 	\begin{subfigure}[t]{0.20\textwidth}
		\centering
		\includegraphics[width=2cm]{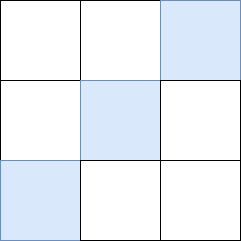}
        \caption*{2}
	\end{subfigure}
  \caption{Samplers used to generate \cref{fig:tri}. Each sampler satisfies the n-rooks property as no two samples taken share the same row or column. Note that when these samplers are randomly applied, all parts of the sampling window have an equal chance of being chosen because each sampler uniquely selects their elements and in total all 9 elements are represented. However, the final sampling is biased to run in diagonals running from the bottom left to top right of the image due to the layout of the samplers. More samplers are required if one wishes to remove this bias (e.g., include the mirror images of these 3 samplers for a total of 6 samplers.}
\label{fig:trisamplers}
\end{figure}

The sequence that generates \cref{fig:tri} using the samplers in \cref{fig:trisamplers}:

0\\
0, 2, 2\\
0, 2, 2, 1, 0, 0, 1, 0, 2\\
1, 1, 0, 0, 2, 1, 1, 1, 2, 1, 2, 1, 1, 1, 2, 0, 2, 0, 1, 2, 0, 0, 0, 0, 0, 1, 2\\

\subsection*{Acknowledgements}
We thank Benjamin Rhoda and David McCarthy of the University of California, Santa Barbara for their discussions and general support as friends.

\bibliographystyle{plain}
\bibliography{my_bib}

\begin{thebibliography}{10}

\bibitem{multipool}
Christian Bailer, Tewodros Habtegebrial, Kiran Varanasi, and Didier Stricker.
\newblock Fast dense feature extraction with cnns that have pooling or striding layers, 09 2017.

\bibitem{deeplab}
Liang{-}Chieh Chen, George Papandreou, Iasonas Kokkinos, Kevin Murphy, and Alan~L. Yuille.
\newblock Deeplab: Semantic image segmentation with deep convolutional nets, atrous convolution, and fully connected crfs.
\newblock {\em CoRR}, abs/1606.00915, 2016.

\bibitem{deeplabv3}
Liang{-}Chieh Chen, George Papandreou, Florian Schroff, and Hartwig Adam.
\newblock Rethinking atrous convolution for semantic image segmentation.
\newblock {\em CoRR}, abs/1706.05587, 2017.

\bibitem{deeplabv3plus2018}
Liang-Chieh Chen, Yukun Zhu, George Papandreou, Florian Schroff, and Hartwig Adam.
\newblock Encoder-decoder with atrous separable convolution for semantic image segmentation.
\newblock {\em arXiv:1802.02611}, 2018.

\bibitem{checkpointing}
Tianqi Chen, Bing Xu, Chiyuan Zhang, and Carlos Guestrin.
\newblock Training deep nets with sublinear memory cost.
\newblock {\em CoRR}, abs/1604.06174, 2016.

\bibitem{deformable}
Jifeng Dai, Haozhi Qi, Yuwen Xiong, Yi~Li, Guodong Zhang, Han Hu, and Yichen Wei.
\newblock Deformable convolutional networks.
\newblock {\em CoRR}, abs/1703.06211, 2017.

\bibitem{fbres}
Facebook.
\newblock Resnet training in torch.
\newblock \url{https://github.com/facebook/fb.resnet.torch}, 2017.

\bibitem{semsegreview}
Alberto Garcia{-}Garcia, Sergio Orts{-}Escolano, Sergiu Oprea, Victor Villena{-}Martinez, and Jos{\'{e}}~Garc{\'{\i}}a Rodr{\'{\i}}guez.
\newblock A review on deep learning techniques applied to semantic segmentation.
\newblock {\em CoRR}, abs/1704.06857, 2017.

\bibitem{fmp}
Benjamin Graham.
\newblock Fractional max-pooling.
\newblock {\em CoRR}, abs/1412.6071, 2014.

\bibitem{pyramidnet}
Dongyoon Han, Jiwhan Kim, and Junmo Kim.
\newblock Deep pyramidal residual networks.
\newblock {\em CoRR}, abs/1610.02915, 2016.

\bibitem{resnet}
Kaiming He, Xiangyu Zhang, Shaoqing Ren, and Jian Sun.
\newblock Deep residual learning for image recognition.
\newblock {\em CoRR}, abs/1512.03385, 2015.

\bibitem{dropout}
Geoffrey~E. Hinton, Nitish Srivastava, Alex Krizhevsky, Ilya Sutskever, and Ruslan Salakhutdinov.
\newblock Improving neural networks by preventing co-adaptation of feature detectors.
\newblock {\em CoRR}, abs/1207.0580, 2012.

\bibitem{squeezeandexcite}
Jie Hu, Li~Shen, and Gang Sun.
\newblock Squeeze-and-excitation networks.
\newblock {\em CoRR}, abs/1709.01507, 2017.

\bibitem{densenet}
Gao Huang, Zhuang Liu, and Kilian~Q. Weinberger.
\newblock Densely connected convolutional networks.
\newblock {\em CoRR}, abs/1608.06993, 2016.

\bibitem{batchnorm}
S.~{Ioffe} and C.~{Szegedy}.
\newblock {Batch Normalization: Accelerating Deep Network Training by Reducing Internal Covariate Shift}.
\newblock {\em ArXiv e-prints}, February 2015.

\bibitem{cifar-pytorch}
kuangliu.
\newblock pytorch-cifar.
\newblock \url{https://github.com/kuangliu/pytorch-cifar}, 2018.

\bibitem{mnist}
Yann LeCun and Corinna Cortes.
\newblock {MNIST} handwritten digit database.
\newblock http://yann.lecun.com/exdb/mnist/.

\bibitem{wide-resnet-pytorch}
meliketoy.
\newblock wide-resnet.pytorch.
\newblock \url{https://github.com/meliketoy/wide-resnet.pytorch}, 2018.

\bibitem{quasi}
A.~B. Owen.
\newblock {Quasi-Monte Carlo} sampling.
\newblock In H.~W. Jensen, editor, {\em Monte Carlo Ray Tracing: Siggraph 2003 Course 44}, pages 69--88. SIGGRAPH, 2003.

\bibitem{memoryefficientdensenet}
Geoff Pleiss, Danlu Chen, Gao Huang, Tongcheng Li, Laurens van~der Maaten, and Kilian~Q Weinberger.
\newblock Memory-efficient implementation of densenets.
\newblock {\em arXiv preprint arXiv:1707.06990}, 2017.

\bibitem{torchvision}
Pytorch.
\newblock Pytorch torchvision.
\newblock \url{https://github.com/pytorch/vision}, 2018.

\bibitem{capsules}
Sara Sabour, Nicholas Frosst, and Geoffrey~E. Hinton.
\newblock Dynamic routing between capsules.
\newblock {\em CoRR}, abs/1710.09829, 2017.

\bibitem{nrooks}
Peter~S. Shirley.
\newblock {\em Physically Based Lighting Calculations for Computer Graphics}.
\newblock PhD thesis, Champaign, IL, USA, 1991.
\newblock UMI Order NO. GAX91-24487.

\bibitem{vgg}
Karen Simonyan and Andrew Zisserman.
\newblock Very deep convolutional networks for large-scale image recognition.
\newblock {\em CoRR}, abs/1409.1556, 2014.

\bibitem{dilated}
Fisher Yu and Vladlen Koltun.
\newblock Multi-scale context aggregation by dilated convolutions.
\newblock {\em CoRR}, abs/1511.07122, 2015.

\bibitem{stochpool}
Matthew~D. Zeiler and Rob Fergus.
\newblock Stochastic pooling for regularization of deep convolutional neural networks.
\newblock {\em CoRR}, abs/1301.3557, 2013.

\bibitem{s3pool}
Shuangfei Zhai, Hui Wu, Abhishek Kumar, Yu~Cheng, Yongxi Lu, Zhongfei Zhang, and Rog{\'{e}}rio~Schmidt Feris.
\newblock S3pool: Pooling with stochastic spatial sampling.
\newblock {\em CoRR}, abs/1611.05138, 2016.

\end{thebibliography}

\end{document}